\documentclass[runningheads]{llncs}
\usepackage{graphicx}
\usepackage{amsmath,amssymb} 
\usepackage[usenames, dvipsnames]{color}
\usepackage[ruled]{algorithm2e}
\usepackage[noend]{algpseudocode}
\usepackage{wrapfig}
\usepackage{tablefootnote}

\newcommand\etal{\textit{et al.}}
\newcommand\ie{\textit{i.e.}}

\usepackage{subfig}
\usepackage[export]{adjustbox}
\usepackage{array}
\usepackage{pbox}
\usepackage{multirow}
\usepackage{booktabs}
\usepackage{hyperref}

\newcolumntype{L}[1]{>{\raggedright\let\newline\\\arraybackslash\hspace{0pt}}m{#1}}
\newcolumntype{C}[1]{>{\centering\let\newline\\\arraybackslash\hspace{0pt}}m{#1}}
\newcolumntype{R}[1]{>{\raggedleft\let\newline\\\arraybackslash\hspace{0pt}}m{#1}}

\begin{document}
\def\ECCV18SubNumber{412}  

\title{Unsupervised Hard Example Mining from Videos for Improved Object Detection} 

\titlerunning{Unsupervised Hard Example Mining from Videos for Object Detection}

\authorrunning{Jin~et~al.}
\newcommand*\samethanks[1][\value{footnote}]{\footnotemark[#1]}

\author{SouYoung Jin\thanks{Authors contributed equally}, Aruni RoyChowdhury\samethanks, Huaizu Jiang, Ashish Singh, \\Aditya Prasad, Deep Chakraborty, and Erik Learned-Miller}

\institute{College of Information and Computer Sciences, University of Massachusetts, Amherst\\
\email{\{souyoungjin,arunirc,hzjiang,ashishsingh,\\aprasad,dchakraborty,elm\}@cs.umass.edu}}

\maketitle

\begin{abstract}
Important gains have recently been obtained in object detection by using training objectives that focus on {\em hard negative} examples, i.e., negative examples that are currently rated as positive or ambiguous by the detector. These examples can strongly influence parameters when the network is trained to correct them. Unfortunately, they are often sparse in the training data, and are expensive to obtain. In this work, we show how large numbers of hard negatives can be obtained {\em automatically} by analyzing the output of a trained detector on video sequences. In particular, detections that are {\em isolated in time}, i.e., that have no associated preceding or following detections, are likely to be hard negatives. We describe simple procedures for mining large numbers of such hard negatives (and also hard {\em positives}) from unlabeled video data. Our experiments show that retraining detectors on these automatically obtained examples often significantly improves performance. We present experiments on multiple architectures and multiple data sets, including face detection, pedestrian detection and other object categories.

\end{abstract}

\section{Introduction}
\label{sec:intro}
\begin{figure}[tb]
\centering
\includegraphics[width=0.8\textwidth]{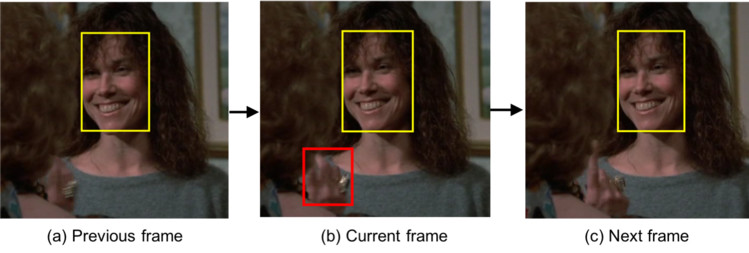}\\
\includegraphics[width=0.8\textwidth]{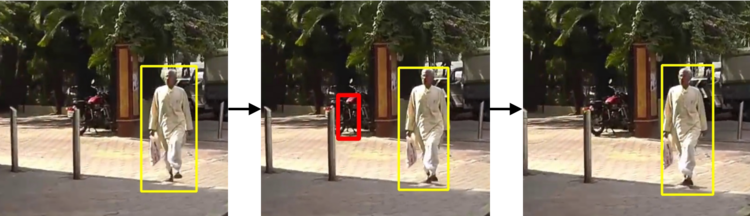}
\caption{\textbf{Detector flicker in videos.} Three consecutive frames from a video are shown for face and pedestrian detection. On the top row, the boxes show face detections from the Faster R-CNN~\cite{ren15faster} (trained on WIDER face)~\cite{yang16wider,jiang2017face}. On the bottom row are detections from the same detector trained on the Caltech pedestrian dataset~\cite{dollar2009pedestrian}. Yellow boxes show true positives and red boxes show false positives. 
For the true positives, the same object is detected in all three frames whereas for the false positives, the detection is \textit{isolated} -- it occurs neither in the previous nor the subsequent frame.  These detections that are ``isolated in time'' frequently turn out to be false positives, and hence  provide important sources of hard negative training data for detectors.}
\label{fig:intro1}
\end{figure}

Detection is a core computer vision problem that has seen major advances in the last few years due to larger training sets, improved architectures, end-to-end training, and improved loss functions~\cite{ren15faster,ren16faster,dollar15fast,zitnick2014edge}. In this work, we consider another direction for improving detectors -- by dramatically expanding the number of hard examples available to the learner. We apply the method to several  different detection problems (including face and pedestrian), a variety of architectures, and multiple data sets, showing significant gains in a variety of settings. 

 Many discriminative methods are more influenced by challenging examples near the boundary of a classifier than easy examples that have low loss. Some classifiers, such as support vector machines, are completely determined by examples near the classifier boundary  (the ``support vectors'')~\cite{scholkopf2002learning}. More recent techniques that emphasize examples near the boundary include general methods such as {\em active bias}~\cite{chang2017active}, which re-weights examples according to the variance of their posteriors during training. In the context of class imbalance in training object detectors, on-line hard example mining (OHEM)~\cite{shrivastava2016training} and the \textit{focal loss}~\cite{lin2017focal} were designed to emphasize hard examples. 

In this paper, we introduce simple methods for automatically mining both hard negatives and hard positives from videos using a previously trained detector. To illustrate, Figure~\ref{fig:intro1} shows a sequence of consecutive video frames from two videos containing a face and a pedestrian respectively. The results of the Faster R-CNN detector (trained for each class) run on each frame are marked as rectangles, with true positives as yellow boxes and false positives as red boxes. 
 Notice that false positives are neither preceded nor followed by a detection. We refer to such isolated-in-time detections as  {\bf detector flickers} and postulate that these are usually caused by false positives rather than true positives.\footnote{Note we are {\it not} claiming that most false positives will be isolated, but only that flickers are likely to be false positives, a very different statement.} This hypothesis stems from the idea that a false positive, caused by something that usually does not look like a face (or other target object), such as a hand, only momentarily causes a detector network to respond positively, but that small deviations from these hard negatives will likely not register as positives. Similar observations can be found in the literature on adversarial examples, where many adversarial examples have been shown to be ``unstable" with respect to minute perturbations of the image~\cite{lu2017no,luo2015foveation,athalye2017synthesizing}. In addition, leveraging the continuity of labelling across space and time has a long history in computer vision. Spatial label dependencies are widely modeled by Markov random fields~\cite{geman1986markov} and conditional random fields~\cite{sutton2006introduction}, while the smoothness of labels across time is a staple of tracking methods and other video processing algorithms~\cite{stalder2010cascaded,klaser2010human,yang2012online}. 

As our experiments show, a large percentage of detector flickers are indeed false positives, and more importantly, they are hard negatives, since they were identified incorrectly as positives by the detector. 
Such an {\it automatically generated training set} of hard negatives can be used to fine-tune a detector, often leading to improved performance. Similar benefits are gained from fine-tuning with \textit{hard positives}, which are obtained in an analogous fashion from cases where a consistently detected object ``flickers off'' in an isolated frame.
While these flickers are relatively rare, it is inexpensive to run a modern detector on many hours of unlabeled video, generating essentially unlimited numbers of hard examples. 
Being an unsupervised process, training sets gathered automatically in this fashion do include some noise. Nevertheless, our experiments show that significant improvements can be gleaned by retraining detectors using these noisy hard examples.
An alternative to gathering such hard examples automatically is, of course, to obtain them manually. However, the rarity of false positives for modern detectors makes this process extremely expensive. Doing this manually requires that every positive detection be examined for validity. With typical false positive rates around one per 1000 images, this process requires the examination of 1000 images per false positive, making it prohibitively expensive.

\section{Related Work}
\label{sec:related}

Convolutional neural networks have recently been applied to achieve state-of-the-art results in object detection~\cite{girshick14rich,girshick15fast,he14spatial,ren16faster,redmon2016you,liu2016ssd,cai2016unified,lin2017feature}.
Many of these object detectors have been re-purposed for other tasks such as face detection~\cite{ranjan15a,li15a,yang15from,farfade15multi}, \cite{li16face,zhang2016joint,yu2016unitbox,jiang2017face,wang2017detecting,hu2017finding,zhang2017s} and pedestrian detection~\cite{zhang2016faster,du2017fused,cai2016unified,cai2015learning,hosang2015taking,li2017scale,zhang2016far}, achieving impressive results~\cite{fddbTech,yang16wider,dollar2009pedestrian}.


\noindent\textbf{Hard negatives in detection.} 
Massive class imbalance is an issue with sliding-window-style object detectors --- being densely applied over an image, such models see far more ``easy'' negative samples from background regions than positive samples from regions containing an object. Some form of hard negative mining is used by most successful object detectors to account for this imbalance~\cite{dalal05histograms,dollar2009integral,felzenszwalb2010object,girshick14rich,girshick15fast,he14spatial,shrivastava2016training,zhang2016faster,lin2017focal,wan2016bootstrapping,sun2017face}.
Early approaches include \textit{bootstrapping}~\cite{sung1994learning} for training SVM-based object detectors~\cite{dalal05histograms,felzenszwalb2010object}, where false positive detections were added to the set of background training samples in an incremental fashion. Other methods~\cite{rowley1998neural,dollar2009integral}  apply a pre-trained detector on a larger dataset to mine false positives and then re-train.

Hard negative mining has also improved the performance of deep learning based models~\cite{simo2014fracking,loshchilov2015online,girshick15fast,shrivastava2016training,zhang2016faster,wan2016bootstrapping,lin2017focal}. 
Shrivastava~\etal~\cite{shrivastava2016training} proposed an \textit{Online Hard Example Mining} (OHEM) procedure,training using only high-loss region proposals. This technique, originally applied to the Fast R-CNN detector~\cite{girshick15fast}, yielded significant gains on the PASCAL and MS-COCO benchmarks.
Lin~\etal~\cite{lin2017focal} propose the \textit{focal loss} to down-weight the contribution of easy examples and train a single-stage, multi-scale network~\cite{lin2017feature}. The A-Fast-RCNN~\cite{wang2017fast} does adversarial generation of hard examples using occlusions and deformations. While similar to our work, our model is trained with hard examples from \textit{real} images and variations are not limited to occlusion and spatial deformations.
Zhang~\etal~\cite{zhang2016faster} show that effective bootstrapping of hard negatives, using a boosted decision forest~\cite{friedman2000additive,appel2013quickly}, significantly improves over a Faster R-CNN baseline for \textit{pedestrian detection}.
Recent \textit{face detection} methods, such as Wan~\etal~\cite{wan2016bootstrapping} and Sun~\etal~\cite{sun2017face}, have also used the bootstrapping of hard negatives to improve the performance of CNN-based detectors --- a pre-trained Faster R-CNN is used to mine hard negatives; then the model is re-trained. However, these methods require a human-annotated dataset of suitable size. Our unsupervised approach does not rely upon bounding-box annotations and thus can be trained upon potentially unlimited data.


\noindent\textbf{Semi-supervised learning.}  Using   mixtures of labeled and unlabeled data is known as \textit{semi-supervised learning}~\cite{blum1998combining,chapelle2009semi,westonlarge}. Rosenberg~\etal~\cite{rosenberg2005semi} ran a trained object detector on unlabeled data and then trained on a subset of this noisy labeled data in an incremental re-training procedure. In Kalal~\etal~\cite{kalal2010pn}, constraints based on video object trajectories are used to correct patch labels of a random forest classifier; these corrected samples are used for re-training. 
Tang~\etal~\cite{tang2012shifting} adapt still-image object detectors to video by selecting training samples from unlabeled videos, based on the consistency between detections and tracklets, and then follow an iterative procedure that selects the easy examples from videos and hard examples from images to re-train the detector. Rather than adapting to the video domain, we seek to improve detector performance on the source domain by selecting hard examples from videos.
Singh~\etal~\cite{singh2016track} gather discriminative regions from weakly-labeled images and then refine their bounding-boxes by incorporating tracking information from weakly-labeled videos. 

\section{Mining Hard Examples from Videos}
\label{sec:method_hnm}

\begin{figure}[!tb]
\centering
\begin{tabular}{  c  c  c  c  }
   	(a)
    &
    \begin{minipage}{.3\textwidth}
      \includegraphics[width=\linewidth]{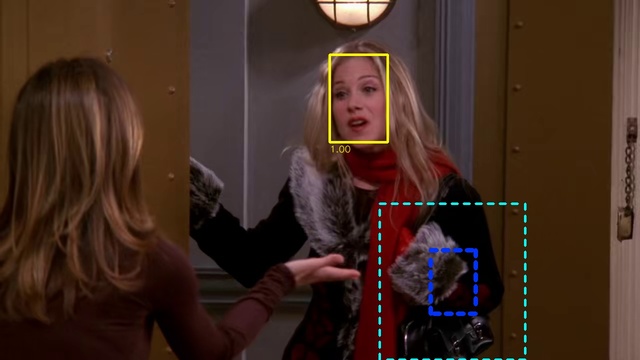}
    \end{minipage}
    &
    \begin{minipage}{.3\textwidth}
      \includegraphics[width=\linewidth]{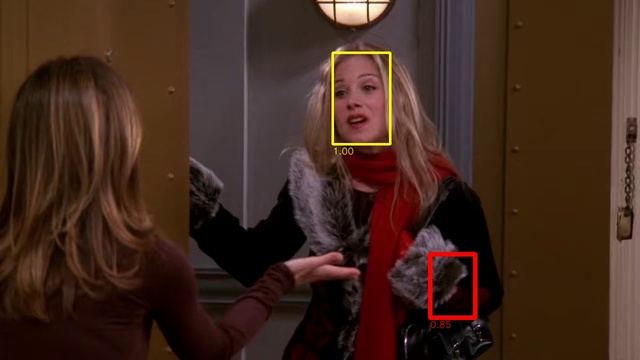}
    \end{minipage}
    &
    \begin{minipage}{.3\textwidth}
      \includegraphics[width=\linewidth]{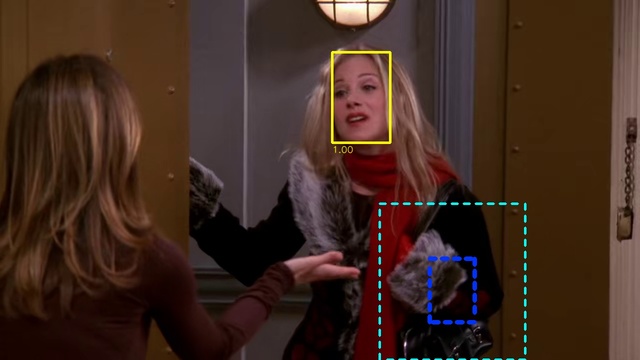}
    \end{minipage}
    \\ 
     \\
    (b)
    &
    \begin{minipage}{.3\textwidth}
      \includegraphics[width=\linewidth]{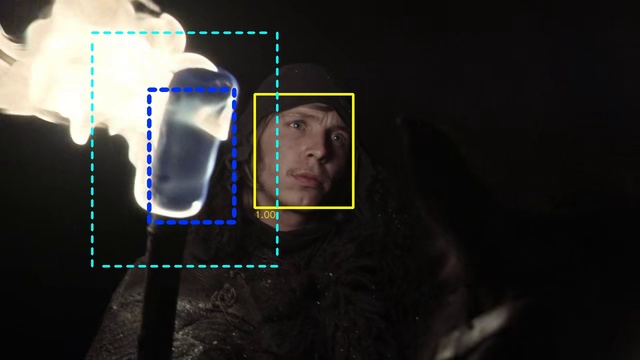}
    \end{minipage}
    &
    \begin{minipage}{.3\textwidth}
      \includegraphics[width=\linewidth]{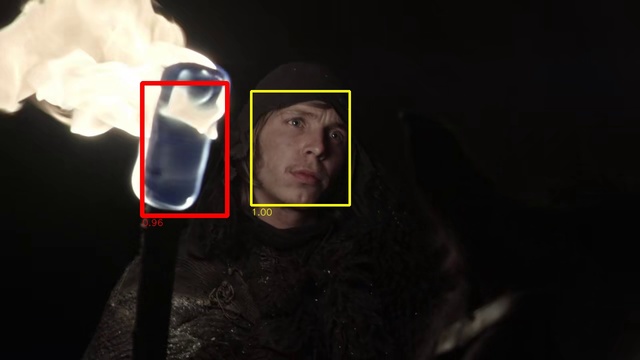}
    \end{minipage}
    &
    \begin{minipage}{.3\textwidth}
      \includegraphics[width=\linewidth]{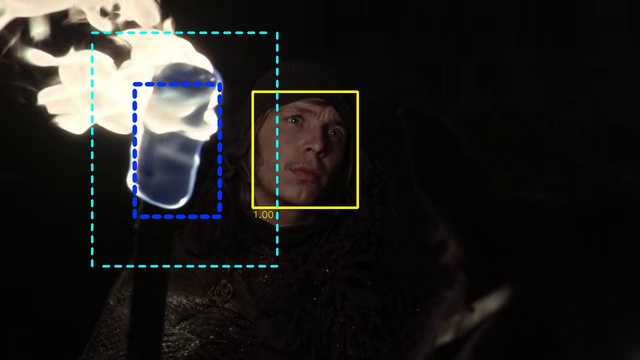}
    \end{minipage}
    \\
    & frame $f$-1 & frame $f$ & frame $f$+1 \\ 
    
  \end{tabular}
  \caption{\textbf{Mining hard negatives from detector-flicker.} The solid boxes denote detections, and the dashed boxes are associated with the tracking algorithm. Given all of the high-confidence \textbf{face detections} in a video (\textbf{\colorbox{yellow}{yellow}} boxes), the proposed algorithm generates a \textbf{tracklet} (\textbf{\colorbox{blue}{\textcolor{white}{blue}}} \textit{dashed} boxes) for the \textbf{current detection} (\textbf{\colorbox{red}{red}} box in frame \textit{f}) by applying template matching within the \textbf{search regions} of the adjacent frames  (\textbf{\colorbox{cyan}{cyan}} \textit{dashed} boxes). As there are no matching detections in adjacent frames for the current detection (\ie~no yellow box matches the blue dashed boxes in frames $f$-1 or $f$+1), it is correctly considered to be an ``isolated detection'' and added to the set of \textbf{\textit{hard negatives}}. The remaining detections in frame $f$, which are temporally consistent, are added to the set of \textbf{\textit{pseudo-positives}}.}
\label{fig:example_face_negative}
\end{figure}

This section discusses methods for automatically mining hard examples from videos, including data collection (Sec.~\ref{sec:data-collect}), our hard negative mining algorithm (Sec.~\ref{sec:mining}), statistics of  recovered hard negatives (Sec.~\ref{sec:stats}) and extension to hard positives (Sec.~\ref{sec:hpm}).
Details of re-training the detector on these new samples are in the Experiments section~(Sec.~\ref{sec:fine-tuning}).

\subsection{Video Collection}
\label{sec:data-collect}
To mine hard examples for face detection, we used 101 videos from sitcoms, each with a duration of 21-25 minutes and a full-length movie of 1 hour 47 minutes, \textit{``Hannah and her sisters''}~\cite{ozerov2013evaluating}. Further, we performed YouTube searches with keywords based on: \textit{public address}, \textit{debate society}, \textit{orchestra performance}, \textit{choir practice} and \textit{courtroom}, downloading 89 videos of durations ranging from 10 to 25 minutes. We obtained videos that were expected to feature a large number of human faces in various scenes, reflecting the everyday settings of our face benchmarks. 
Similarly, for pedestrian detection, we collected videos from YouTube by searching with the two key phrases: {\em driving cam videos} and {\em walking videos}. We obtained 40 videos with an average duration of about 30 minutes.


\subsection{Hard Negative Mining}
\label{sec:mining}

Running a pre-trained face detector on every frame of a video gives us a large set of detections with noisy labels. We crucially differ here from recent bootstrapping approaches~\cite{wan2016bootstrapping,sun2017face} by (a) using large amounts of \textit{unlabeled} data available on the web instead of relying only on the limited fully-supervised  training data from WIDER Face~\cite{yang16wider} or Caltech Pedestrians~\cite{dollar2009pedestrian}, and  (b) having a novel filtering criterion on the noisy labels obtained from the detector that retains the hard negative examples and minimizes noise in the obtained labels. 



The raw detections from a video were thresholded at a relatively high confidence score of 0.8. For every detection in a frame, we formed a short tracklet by performing template matching in adjacent frames, within a window of $\pm 5$ frames --- the bounding box of the current detection was enlarged by 100 pixels and this region was searched in adjacent frames for the best match using normalized cross correlation (NCC). 
To account for occlusions, we put a threshold on the NCC similarity score (set as 0.5) to reject cases where there was a lot of appearance-change between frames. Now in each frame, if the maximum intersection-over-union (IoU) between the tracklet prediction and detections in the adjacent frames was below 0.2, we considered it to be an isolated detection resulting from \textbf{detector flicker}. These isolated detections were taken as \textbf{\textit{hard negatives}}. The detections that \textit{were} found to be consistent with adjacent frames were considered to have a high probability of being true predictions and were termed \textbf{\textit{pseudo-positives}}. For the purpose of creating the re-training set, we kept only those frames that had at least one pseudo-positive detection in addition to one or more hard negatives. Illustrative examples of this procedure are shown in Figure~\ref{fig:example_face_negative}, where we visualize only the previous and next frames for simplicity.

\subsection{Results of Automatic Hard Negative Mining}
\label{sec:stats} 

Our initial mining experiments were performed using a standard Faster R-CNN detector trained on WIDER Face~\cite{yang16wider} for faces and Caltech~\cite{dollar2009pedestrian} for pedestrians.
We collected 13,888 video frames for faces, where each frame contains at least one pseudo-positive and one hard negative (detector flicker). To verify the quality of our automatically mined hard negatives, we randomly sampled 511 hard negatives for inspection. 453 of them are true negatives, while 16 samples are true positives, and 42 samples are categorized as {\em ambiguous}, which correspond extreme head pose or severe occlusions. The precision for true negatives is 88.65\% and precision for true negatives plus {\em ambiguous} is 96.87\%. 

For pedestrians, we collected 14,967 video frames. We manually checked 328 automatically mined hard negatives, where 244 of them are true negatives and 21 belong to \emph{ambiguous}. The precision for true negatives is 74.48\% and precision for true negatives plus \emph{ambiguous} is 82.18\%. 

To further validate our method on an existing fully-annotated video dataset, we used the Hannah dataset~\cite{ozerov2013evaluating}, which has every frame annotated with face bounding boxes. Here, out of 234 mined hard negatives, 187 were true negatives, resulting in a precision of 79.91\%. We note that the annotations on the Hannah movie are not always consistent and involve a significant domain shift from WIDER. Considering the fact no human supervision is provided, the mined face hard negatives are consistently of high quality across various domains.

\subsection{Extension to Hard Positive Mining}
\label{sec:hpm}
\begin{figure}[tb]
\centering
\begin{tabular}{@{\extracolsep{2pt}}ccccc}
\includegraphics[width=0.19\linewidth]{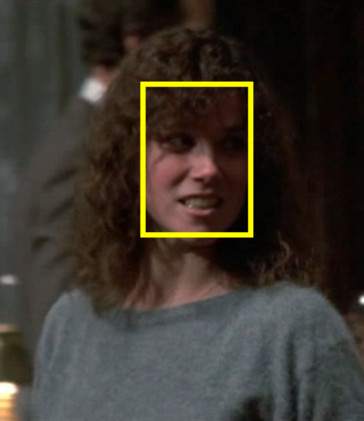} &
\includegraphics[width=0.19\linewidth]{image/hp2} &
\includegraphics[width=0.19\linewidth]{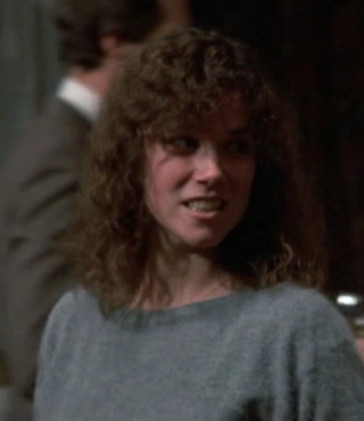} &
\includegraphics[width=0.19\linewidth]{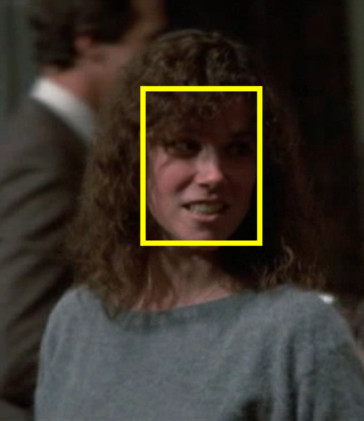} &
\includegraphics[width=0.19\linewidth]{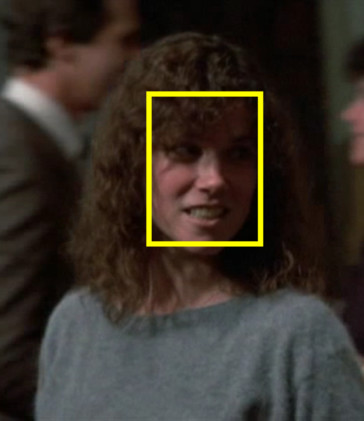} \\
frame $f$-2 & frame $f$-1 & frame $f$ & frame $f$+1 & frame $f$+2\\
\end{tabular}
\caption{{\bf Hard positive samples.} Given a sequence of video frames, the face of the actor is consistently detected except at frame $f$. Such isolated ``off-flickers'' can be harvested in an unsupervised fashion to form a set of \textit{hard positives}. }
\label{fig:hardpositive}
\end{figure}

In principle, the same concept for using detector flickers can be directly applied to obtaining \textbf{\textit{hard positives}}. The idea is to look for ``off-flickers" of a detector in a video tracklet -- given a series of detections of an object in a video, such as a face, we can search for single frames that have no detections but are surrounded by detections on either side. Of course, these could be caused by short-duration occlusions, for example, but a large percentages of these ``off-flickers'' are hard positives, as in Fig.~\ref{fig:hardpositive}. We generate tracklets using the method from~\cite{erdosrenyi} and show results incorporating hard positives on pedestrian and face detection in the experiments section. 
The manually calculated purity over 300 randomly sampled frames was 94.46\% for faces and 83.13\% for pedestrians.


\section{Experiments}
\label{sec:experiment}



We evaluate our method on face and pedestrian detection and perform ablation studies analyzing the effect of the hard examples.For pedestrians, we show results on the Caltech dataset~\cite{dollar2009pedestrian}, while for face detection, we show results on the WIDER Face~\cite{yang16wider} dataset.

The Caltech Pedestrian Dataset~\cite{dollar2009pedestrian} consists of videos taken from a vehicle driving through urban traffic, with about 350k annotated bounding-boxes from 250k video frames. 

The WIDER dataset consists of 32,203 images having 393,703 labeled faces in challenging situations of scale, pose and occlusion. The evaluation set of WIDER is divided into \textit{easy}, \textit{medium}, and \textit{hard} sets according to the detection scores of object proposals from EdgeBox~\cite{zitnick2014edge}. From easy to hard, the faces get smaller and more crowded. 
\subsection{Retraining Detectors with Mined Hard Examples} 
\label{sec:fine-tuning}

We experimented with two ways to leverage our mined \textit{hard negative} samples. In our initial experiments, a single mini-batch is formed by including one image from the original labeled training dataset and another image sampled from our automatically-mined hard negative video frames. In this way, positive region proposals are sampled from the original training dataset image, based on manual annotation, while negative region proposals are sampled from both the original dataset image and the mined hard negative video frame. Thus, we can \textit{explicitly} force the network to focus on the hard negatives from the mined video frame. However, this method did not produce better results in our initial experiments. An alternate approach was found to be more effective -- we simply provided the \textit{pseudo-positives} in the mined video frames as true object annotations during training and \textit{implicitly} allowed the network to pick the hard-negatives.   The inclusion of video frames with \textit{hard positives} is more straightforward -- we can simply treat them as additional images with object annotations at training time. The models were fine-tuned with and without OHEM, and we consistently chose the setting that gave the best validation results. While OHEM would increase the likelihood of hard negatives being selected in a mini-batch, it would also place extra emphasis on any mislabels in the hard examples. This would magnify the effect of a small amount of label noise and can in some cases decrease the overall performance. 


\subsection{Ablation Settings}
\label{sec:ablation_study}
In addition to the comparisons to the baseline Faster R-CNN detectors, we conduct various ablation studies on the Caltech Pedestrian and WIDER Face datasets to address the effectiveness of hard example mining. 

\noindent\textbf{Effect of training iterations.} To account for the possible situation where simply training the baseline model longer may result in a gain in performance, we create another baseline by fine-tuning the original model for additional iterations with a lower learning rate, matching the number of training iterations used in our hard example trained models. We refer to this model as ``\texttt{w/ more iterations}''.

\noindent\textbf{Effect of additional video frames.} Unlike the baseline detector, our fine-tuned models use additional video frames for training. It's possible that just using the high-confidence detection results on unlabeled video frames as \textit{pseudo-groundtruths} during training is sufficient to boost performance, without correcting the hard negatives using our detector flicker approach. 
Therefore we train another detector, ``\texttt{Flickers as Positives}'', starting from the baseline model, that takes exactly the same training set as our hard negative model, but where \textit{all} the high-confidence detections on the video frames are used as positive labels. 


\noindent\textbf{Effect of automatically mined hard examples.} We include the results from our proposed method of considering detector flickers as hard negatives and hard positives separately -- ``\texttt{Flickers as HN}'' and ``\texttt{Flickers as HP}''. Finally, we report results from fine-tuning the detector on the union of both types of hard examples (\texttt{Flickers as HN + HP}).


\subsection{Pedestrian Detection}
\label{sec:pedes}

For our \texttt{baseline} model, we train the VGG16-based \textit{\textbf{Faster R-CNN}} object detector~\cite{ren15faster} with OHEM~\cite{shrivastava2016training} for 150K iterations on the \textbf{Caltech Pedestrian} training dataset~\cite{dollar2009pedestrian}. We used \textit{all} the frames from set00-set05 (which constitute the training set), irrespective of whether they are flagged as ``reasonable'' or not by the Caltech meta-data. Following Zhang~\etal~\cite{zhang2016faster}, we set the IoU ratio for RPN training to 0.5, while all the other experimental settings are identical to~\cite{ren15faster}. The number of labeled Caltech images is 128,419 and our mining provides 14,967 hard negative and 42,914 hard positive frames.
We fine-tune the baseline model with hard examples and the annotated examples from the Caltech Pedestrian {\em training} dataset, with a fixed learning rate of 0.0001 for 60K iterations, using OHEM.
We evaluate our model on the Caltech Pedestrian testing dataset under the {\em reasonable} condition.

\begin{figure}
\centering
\begin{tabular} {@{\extracolsep{2pt}}cc}
\scalebox{1}{\includegraphics[width=0.45\textwidth]{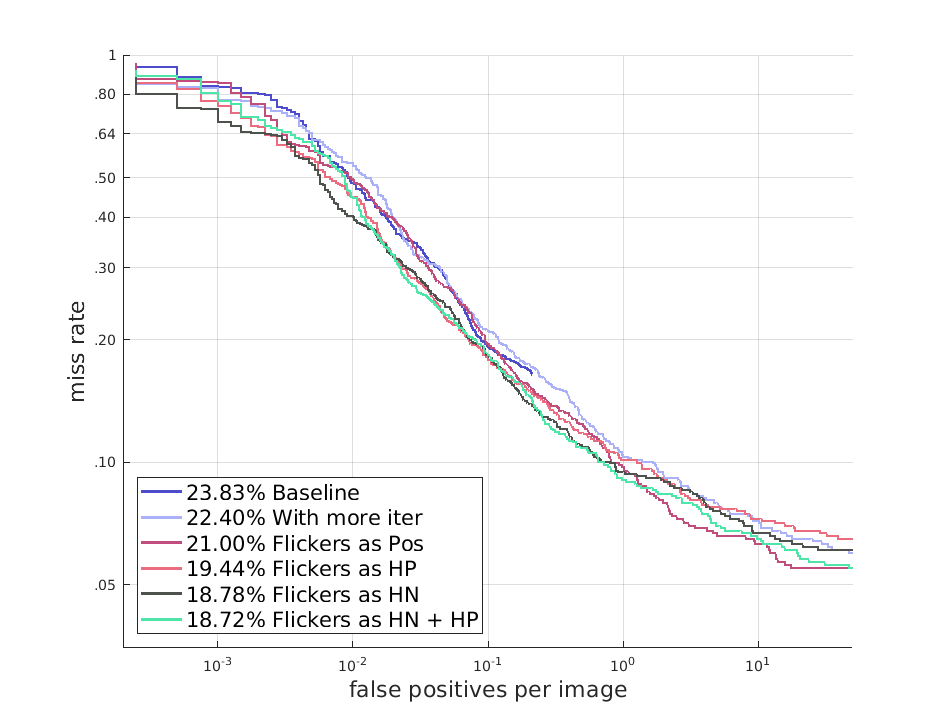}} &
\scalebox{1}{\includegraphics[width=0.45\textwidth]{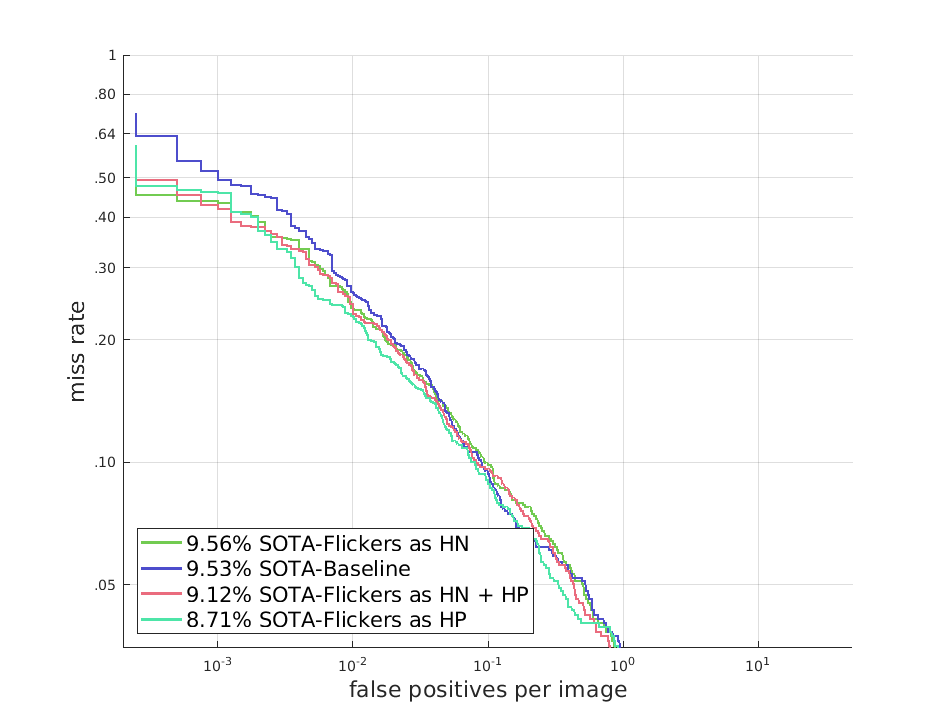}} \\
(a) & (b) \\
\end{tabular}
\caption{ Results on the  \textbf{Caltech Pedestrian} dataset~\cite{dollar2009pedestrian} in {\em reasonable} condition. (a) Faster R-CNN results: using hard negative samples (\texttt{Flickers as HN}) and hard positive samples (\texttt{Flickers as HP}) improve the performance over the baseline in; using a combination of both gives the best performance. (b) State-of-the-art SDS-RCNN results: \texttt{Flickers as HN} improves the original SDS-RCNN results only in the low false positive regime, while \texttt{Flickers as HP} gives the best results. }
\label{fig:caltech_curves}
\end{figure}

The ROC curves of various settings of our models are shown in Fig.~\ref{fig:caltech_curves}(a).
Fine-tuning the existing detector for more iterations gives a modest reduction in log average miss rate, from 23.83\% to 22.4\%. Using all detections without correcting the hard negatives (\texttt{Flickers as Pos}) also gives a small improvement -- the extra training data, although noisy, still has some positive contribution during fine-tuning.
Our proposed model, fine-tuned with the mined hard negatives (\texttt{Flickers as HN}), has a log average miss rate of \textbf{18.78\%}, which outperforms the \texttt{baseline model} by \textbf{5.05\%}. Fine-tuning with hard positives (\texttt{Flickers as HP}) also shows an improvement of \textbf{4.39\%} over the baseline. Combining both hard positives and hard negatives results in the best performance of \textbf{18.72\%} log average miss rate.


In Figure~\ref{fig:caltech_curves}(b) we report results using the state-of-the-art \textbf{\textit{SDS-RCNN}}~\cite{brazil2017illuminating} pedestrian detector~\footnote{Running the authors' released code from \url{https://github.com/garrickbrazil/SDS-RCNN}}. Every 3rd frame is sampled from the Caltech dataset for training the original detector~\cite{brazil2017illuminating}, and we keep this setting in our experiments. For SDS-RCNN, there are 42,782 labeled training images while the mining gives us 2,191 hard negative and 177,563 hard positive frames. The inclusion of hard negatives in training (\texttt{Flickers as HN}) improves the performance of SDS-RCNN in the low False Positives regime compared to the baseline -- the detector learns to eliminate a number of false detections, thereby increasing precision, but it also ends up hurting the recall. Including mined hard positives (\texttt{Flickers as HP}) we get the best performance of \textbf{8.71\%} log average miss rate, outperforming the model using both the mined hard negative and positive samples (\texttt{Flickers as HP + HN}), which gets 9.12\%.

\subsection{Face Detection}
\label{sec:face_wider}

 We adopt the Faster R-CNN framework, using VGG16 as the backbone network. We first train a baseline detector starting from an ImageNet pre-trained model, with a fixed learning rate of 0.001 for 80K iterations using the SGD optimizer, where the momentum is 0.9 and weight decay is 0.0005. 
 For hard negatives, the model is fine-tuned for 50k iterations with learning rate 0.0001. For hard positives, and the combination of both types of hard examples, we train longer for 150k iterations.  Following the \textbf{WIDER Face} protocol, we report Average Precision (AP) values in Table~\ref{tab:wider_res} on the three splits -- `Easy', `Medium' and `Hard'. OHEM is not used as it was empirically observed to decrease performance.

Fine-tuning the baseline model for more iterations improves performance slightly on the Easy and Medium splits. Naively considering all the high confidence detections as true positives (\texttt{Flickers as Positives}) degrades performance substantially across all splits. Hard negative mining, \texttt{Flickers as HN}, slightly outperforms the baseline Faster R-CNN detector (\texttt{w/ more iterations}) on the Medium and Hard splits, retaining the same performance of 0.907 AP on the Easy split. Using the mined hard positives, \texttt{Flickers as HP}, we observe a significant gain in performance on all three splits. Using both hard positives and hard negatives jointly (\texttt{Flickers as HP + HN}) improves over using hard negatives and the baseline, but the improvement is less than the gains from \texttt{Flickers as HP}.

For faces, we additionally experimented with the recent RetinaNet~\cite{lin2017focal} detector as a second high-performance baseline model. Unfortunately, inclusion of the unlabeled data hurt performance slightly using this model, despite the reasonably high purity of the mined examples. While the purity of our mined examples is high, it is not perfect. These incorrect samples would be strongly emphasized by the focal loss used in RetinaNet. Thus, it is possible that while RetinaNet outperforms the Faster R-CNN on standard benchmarks, it may be more susceptible to label noise and thus not a good candidate for our method. In the future, we will investigate different values of the focal loss parameter to see whether this can mitigate the effects of label noise.


\begin{table}[htbp]
\renewcommand{\tabcolsep}{5pt}
\centering
\caption{Average precision (AP) on the validation set of the \textbf{WIDER Face}~\cite{yang16wider} benchmark. Including hard examples improves performance over the baseline, with \texttt{HP} and \texttt{HP+HN} giving the best results.}
\label{tab:wider_res}
\begin{tabular}{@{\extracolsep{5pt}}clccc}
\toprule
 & & Easy & Medium & Hard \\
\midrule
\multirow{6}{*}{Faster R-CNN} & Baseline & 0.907 & 0.850 & 0.492 \\
                            & w/ more iterations & 0.910 & 0.852 & 0.493 \\
                            & Flickers as Positives & 0.829 & 0.790 & 0.434 \\
                            & \textbf{Ours:} Flickers as HN & 0.909 & 0.853 & 0.494 \\
                            & \textbf{Ours:} Flickers as HP & \textbf{0.921} & \textbf{0.864} & 0.492 \\
                            & \textbf{Ours:} Flickers as HP + HN & \textbf{0.921} & \textbf{0.864} & \textbf{0.497} \\
\bottomrule
\end{tabular}
\end{table}




\begin{figure}[t]
\centering
\begin{tabular}{@{\extracolsep{2pt}}ccc}
\includegraphics[width=0.32\linewidth]{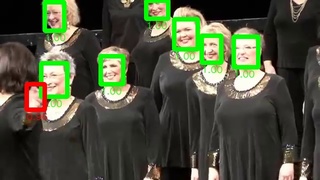} &
\includegraphics[width=0.32\linewidth]{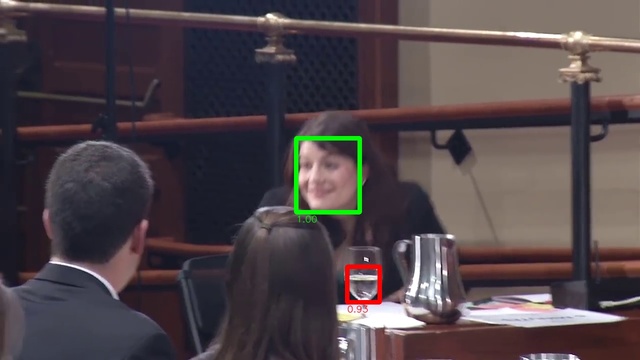} &
\includegraphics[width=0.32\linewidth]{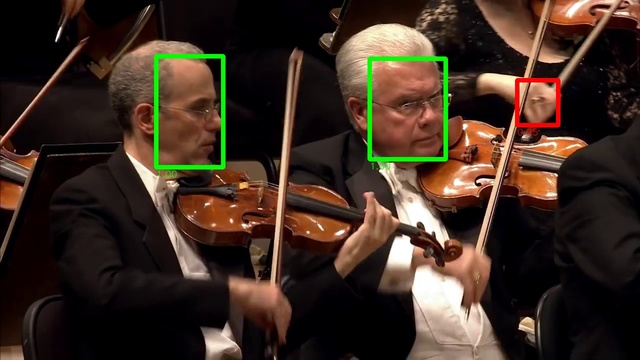} \\
\includegraphics[width=0.32\linewidth]{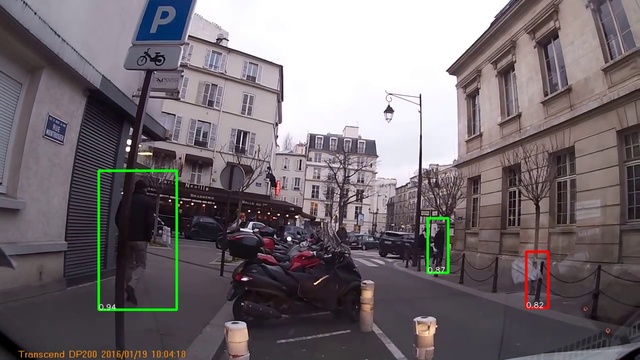} &
\includegraphics[width=0.32\linewidth]{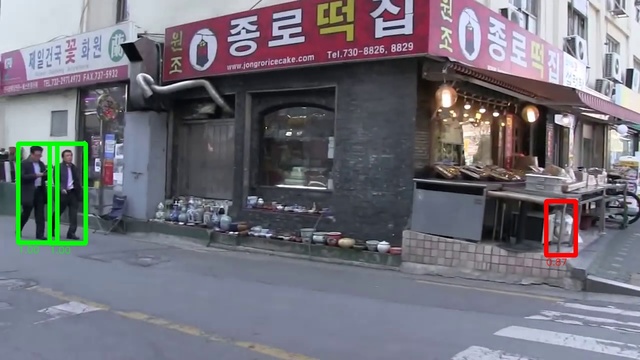} &
\includegraphics[width=0.32\linewidth]{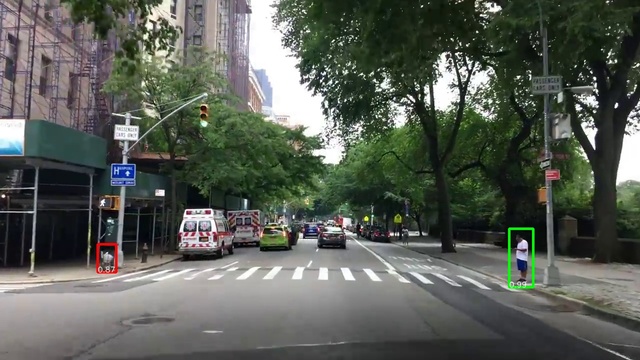} \\
\end{tabular}
\caption{{\bf Examples of hard negatives.} Visualization of  mined hard negatives for faces (\textit{top row}) and pedestrians (\textit{bottom row}). Red boxes denote the ``detection-flicker cases'' among the high confidence detections (green boxes).}
\label{fig:hardnegatives}
\end{figure}



\begin{figure}
\centering
\begin{tabular}{c c  c  c  c  c  c }
    & Groundtruth & Baseline & HN & HP & HP+HN \\
    F1
    &
    \begin{minipage}{.18\textwidth}
      \includegraphics[width=\linewidth]{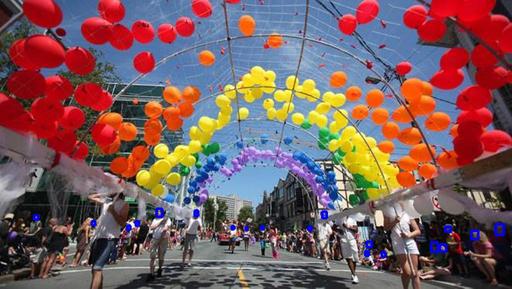}
    \end{minipage}
    &
    \begin{minipage}{.18\textwidth}
      \includegraphics[width=\linewidth]{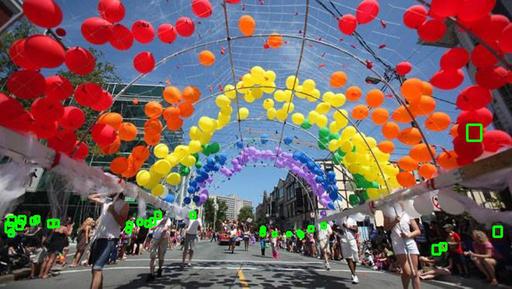}
    \end{minipage}
    &
    \begin{minipage}{.18\textwidth}
      \includegraphics[width=\linewidth]{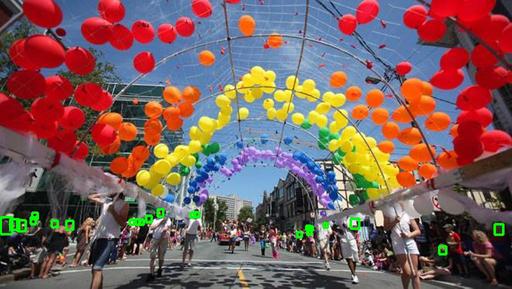}
    \end{minipage}
    &
    \begin{minipage}{.18\textwidth}
      \includegraphics[width=\linewidth]{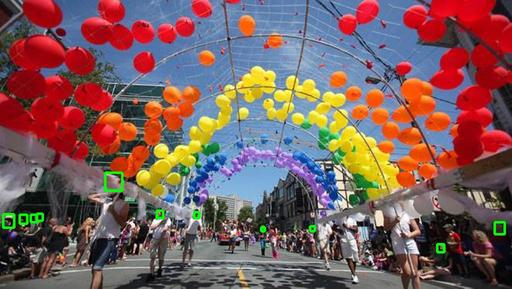}
    \end{minipage}
    &
    \begin{minipage}{.18\textwidth}
      \includegraphics[width=\linewidth]{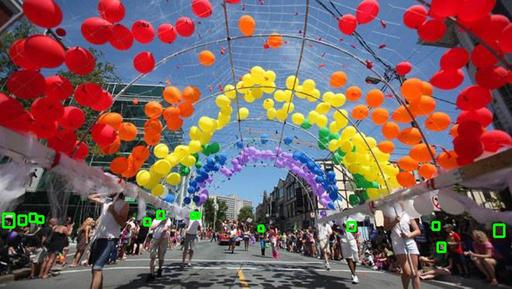}
    \end{minipage}
    \\ 
    F2
    &
    \begin{minipage}{.18\textwidth}
      \includegraphics[width=\linewidth]{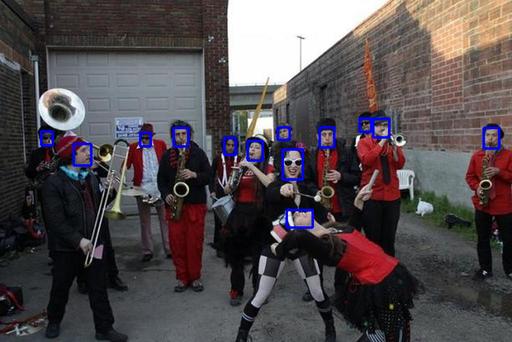}
    \end{minipage}
    &
    \begin{minipage}{.18\textwidth}
      \includegraphics[width=\linewidth]{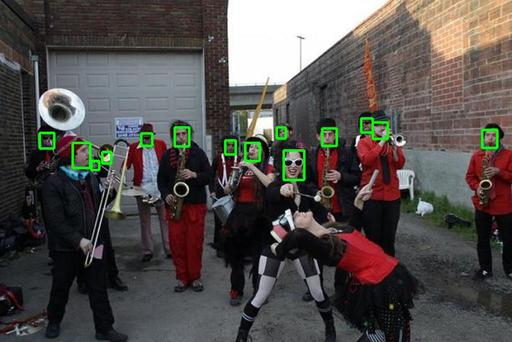}
    \end{minipage}
    &
    \begin{minipage}{.18\textwidth}
      \includegraphics[width=\linewidth]{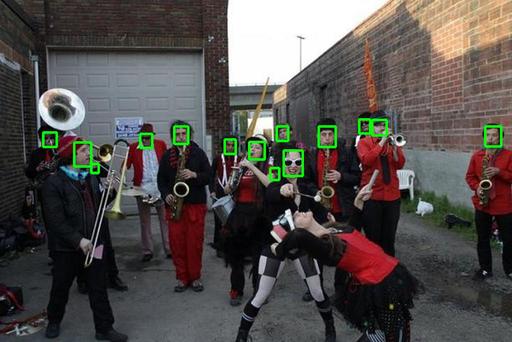}
    \end{minipage}
    &
    \begin{minipage}{.18\textwidth}
      \includegraphics[width=\linewidth]{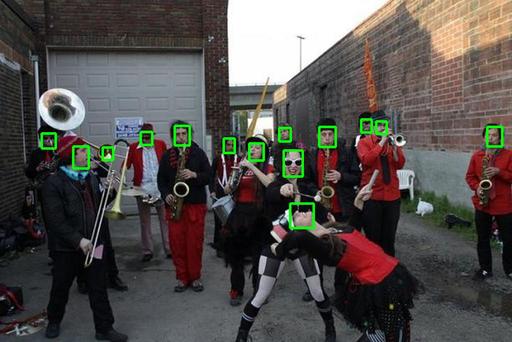}
    \end{minipage}
    &
    \begin{minipage}{.18\textwidth}
      \includegraphics[width=\linewidth]{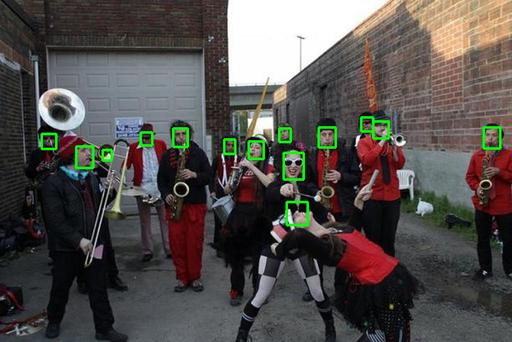}
    \end{minipage}
    \\ 
    F3
    &
    \begin{minipage}{.18\textwidth}
      \includegraphics[width=\linewidth]{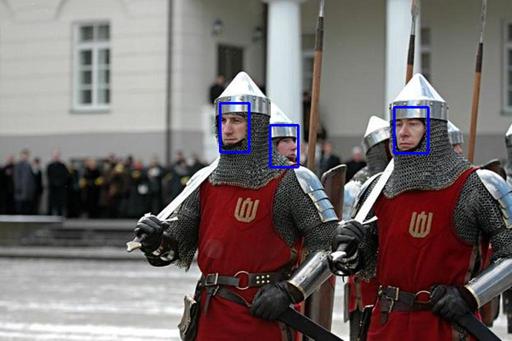}
    \end{minipage}
    &
    \begin{minipage}{.18\textwidth}
      \includegraphics[width=\linewidth]{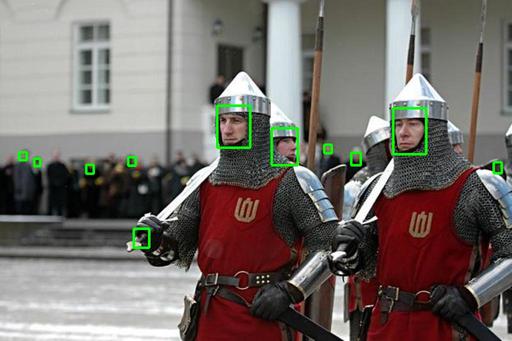}
    \end{minipage}
    &
    \begin{minipage}{.18\textwidth}
      \includegraphics[width=\linewidth]{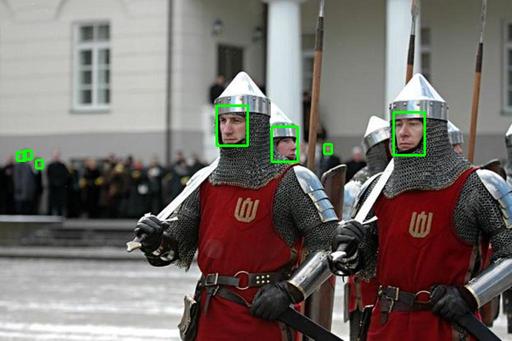}
    \end{minipage}
    &
    \begin{minipage}{.18\textwidth}
      \includegraphics[width=\linewidth]{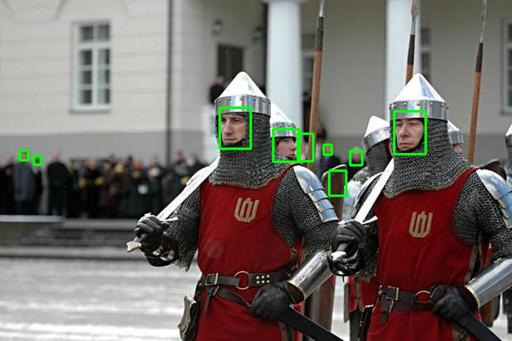}
    \end{minipage}
    &
    \begin{minipage}{.18\textwidth}
      \includegraphics[width=\linewidth]{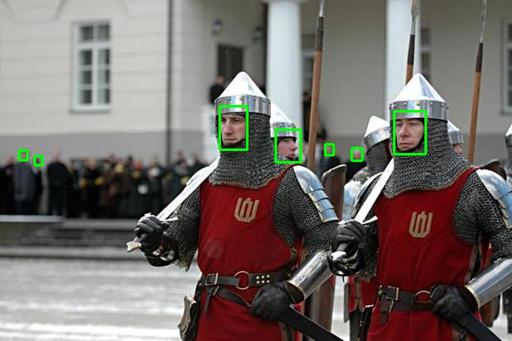}
    \end{minipage}
    \\
    F4
    &
    \begin{minipage}{.18\textwidth}
      \includegraphics[width=\linewidth]{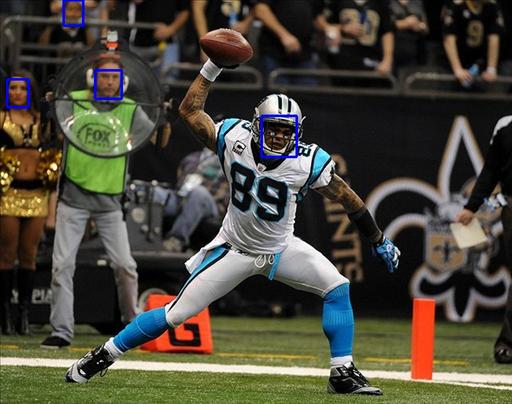}
    \end{minipage}
    &
    \begin{minipage}{.18\textwidth}
      \includegraphics[width=\linewidth]{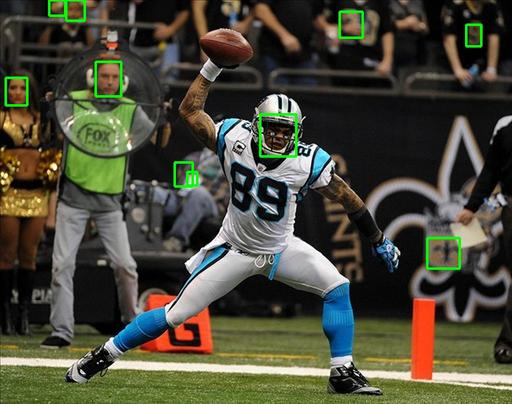}
    \end{minipage}
    &
    \begin{minipage}{.18\textwidth}
      \includegraphics[width=\linewidth]{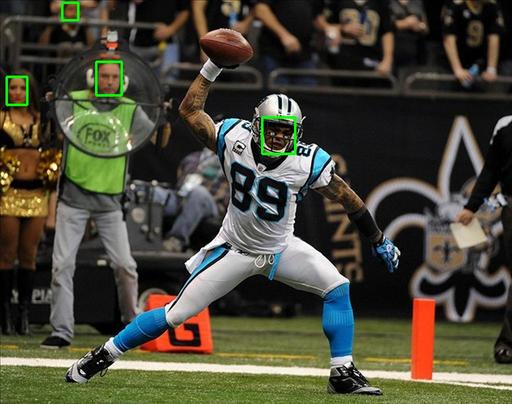}
    \end{minipage}
    &
    \begin{minipage}{.18\textwidth}
      \includegraphics[width=\linewidth]{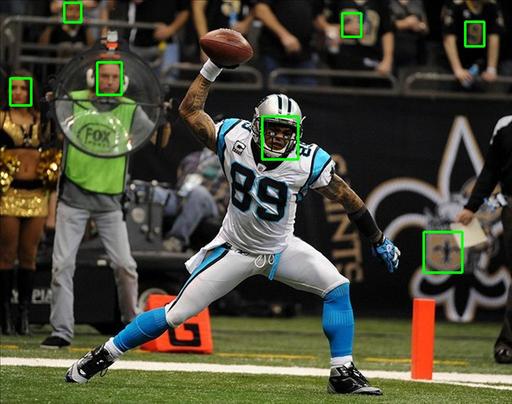}
    \end{minipage}
    &
    \begin{minipage}{.18\textwidth}
      \includegraphics[width=\linewidth]{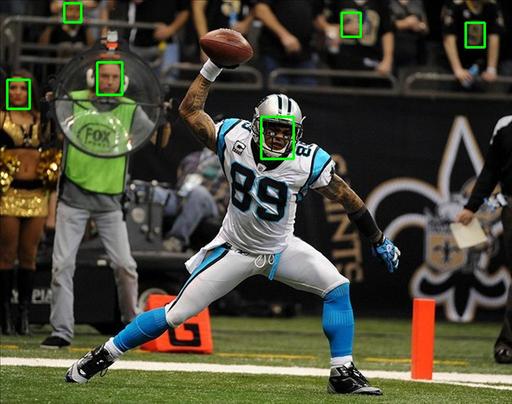}
    \end{minipage}
    \\ 
    \\
    P1
    &
    \begin{minipage}{.18\textwidth}
      \includegraphics[width=\linewidth]{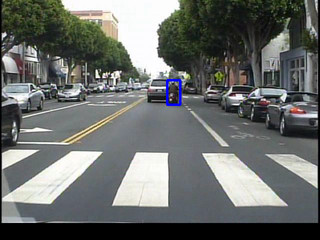}
    \end{minipage}
    &
    \begin{minipage}{.18\textwidth}
      \includegraphics[width=\linewidth]{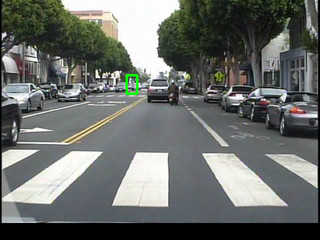}
    \end{minipage}
    &
    \begin{minipage}{.18\textwidth}
      \includegraphics[width=\linewidth]{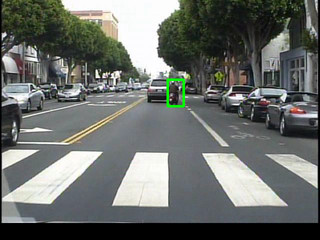}
    \end{minipage}
    &
    \begin{minipage}{.18\textwidth}
      \includegraphics[width=\linewidth]{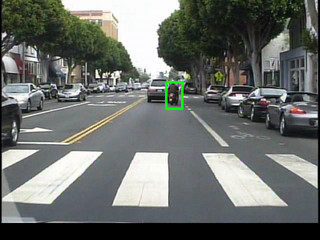}
    \end{minipage}
    &
    \begin{minipage}{.18\textwidth}
      \includegraphics[width=\linewidth]{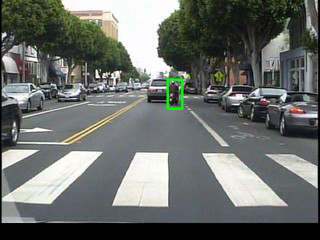}
    \end{minipage}
    \\ 
    P2
    &
    \begin{minipage}{.18\textwidth}
      \includegraphics[width=\linewidth]{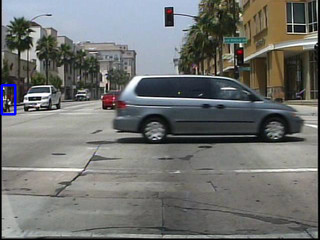}
    \end{minipage}
    &
    \begin{minipage}{.18\textwidth}
      \includegraphics[width=\linewidth]{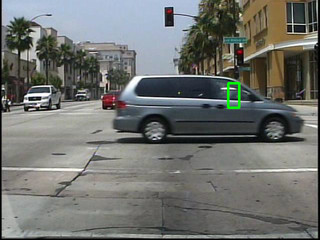}
    \end{minipage}
    &
    \begin{minipage}{.18\textwidth}
      \includegraphics[width=\linewidth]{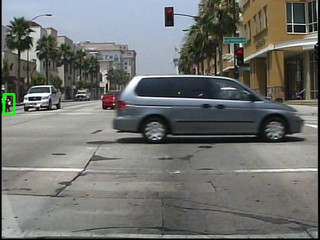}
    \end{minipage}
    &
    \begin{minipage}{.18\textwidth}
      \includegraphics[width=\linewidth]{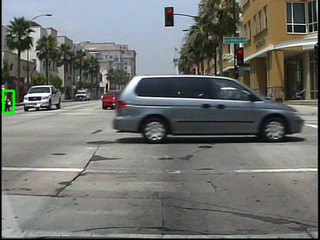}
    \end{minipage}
    &
    \begin{minipage}{.18\textwidth}
      \includegraphics[width=\linewidth]{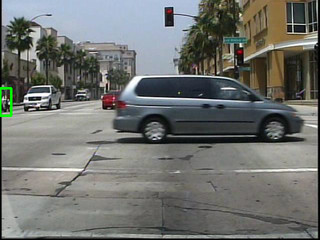}
    \end{minipage}
    \\ 
    P3
    &
    \begin{minipage}{.18\textwidth}
      \includegraphics[width=\linewidth]{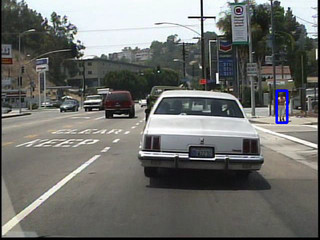}
    \end{minipage}
    &
    \begin{minipage}{.18\textwidth}
      \includegraphics[width=\linewidth]{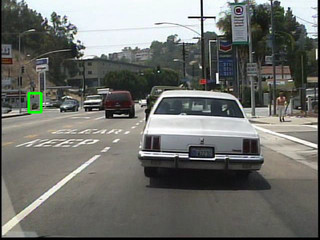}
    \end{minipage}
    &
    \begin{minipage}{.18\textwidth}
      \includegraphics[width=\linewidth]{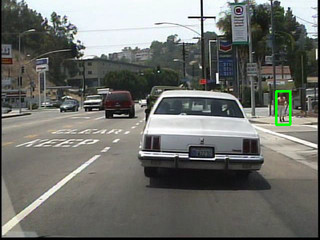}
    \end{minipage}
    &
    \begin{minipage}{.18\textwidth}
      \includegraphics[width=\linewidth]{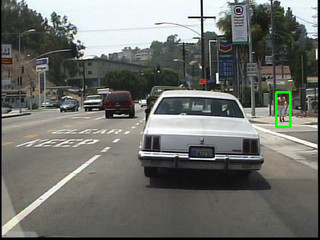}
    \end{minipage}
    &
    \begin{minipage}{.18\textwidth}
      \includegraphics[width=\linewidth]{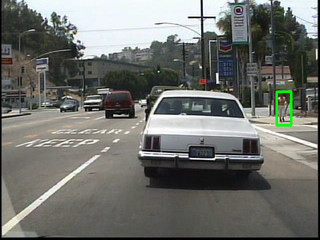}
    \end{minipage}
    \\
    P4
    &
    \begin{minipage}{.18\textwidth}
      \includegraphics[width=\linewidth]{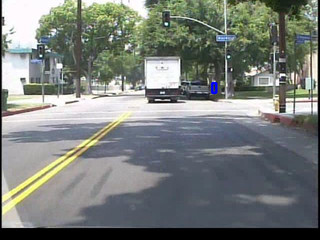}
    \end{minipage}
    &
    \begin{minipage}{.18\textwidth}
      \includegraphics[width=\linewidth]{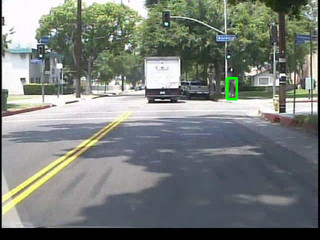}
    \end{minipage}
    &
    \begin{minipage}{.18\textwidth}
      \includegraphics[width=\linewidth]{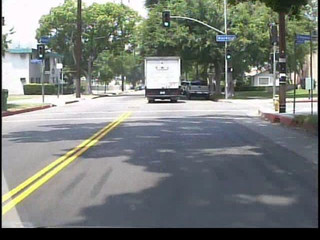}
    \end{minipage}
    &
    \begin{minipage}{.18\textwidth}
      \includegraphics[width=\linewidth]{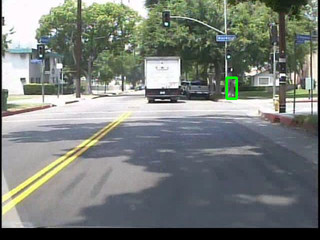}
    \end{minipage}
    &
    \begin{minipage}{.18\textwidth}
      \includegraphics[width=\linewidth]{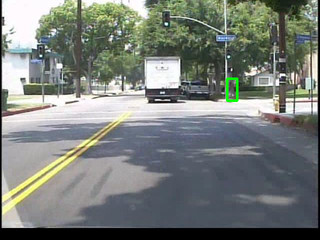}
    \end{minipage}
    \\
    
  \end{tabular}
  \caption{\textbf{Qualitative comparison.} Faster R-CNN detections for faces (F1-4) and pedestrians (P1-4).The detector fine-tuned with hard negatives (HN) reduces false positives compared to the Baseline (F-1,3,4; P-1,2,3), but can sometimes lower the recall (P4). Hard positives (HP) increases recall (F2, P4) but can also introduce false positives (F4). Using both (HP+HN) the detector is usually able to achieve a good balance.}
\label{fig:vis_dets_hn_hp}
\end{figure}

\section{Discussion}
\label{sec:discussion}

In this section, we discuss some further applications and extensions to our proposed hard example mining method.

\noindent\textbf{On the Entropy of the False Positive Distribution.}
In mining thousands of hard negatives from unlabeled video, we noticed a striking pattern in the hard negatives of face detectors. A large percentage of false positives were generated by a few types of objects. Specifically, a large percentage of hard negatives in face detectors seem to stem from human hands, ears, and the torso/chest area. 
Since it appears that a large percentage of the false positives in face detection are the result of a relatively small number of phenomena, this could explain the significant gains realized by modeling hard negatives. In particular, characterizing the distribution of hard negatives, and learning to avoid them, may involve a relatively small set of hard negatives.

\begin{figure}[h]
\centering
\begin{tabular} {@{\extracolsep{2pt}}ccc}
\scalebox{1}{\includegraphics[width=0.3\textwidth]{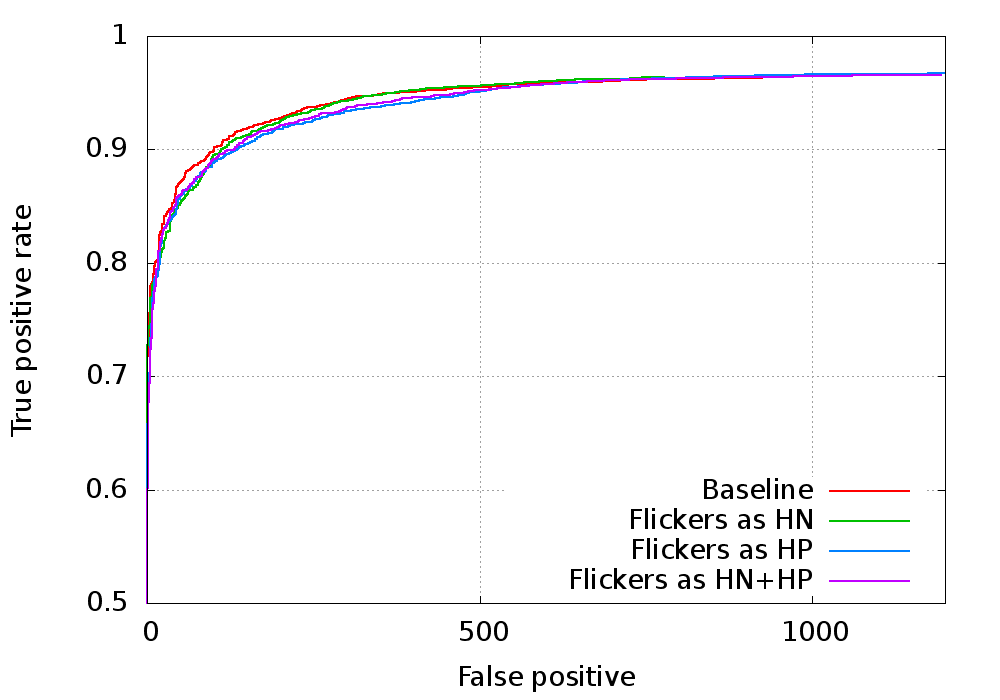}} &
\scalebox{1}{\includegraphics[width=0.3\textwidth]{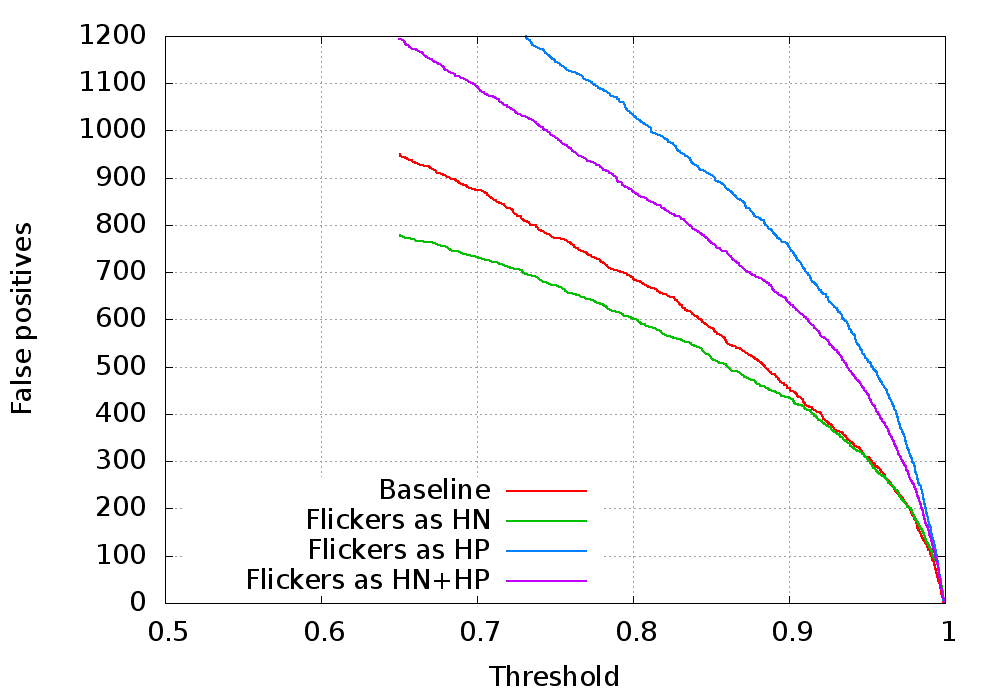}} & 
\scalebox{1}{\includegraphics[width=0.3\textwidth]{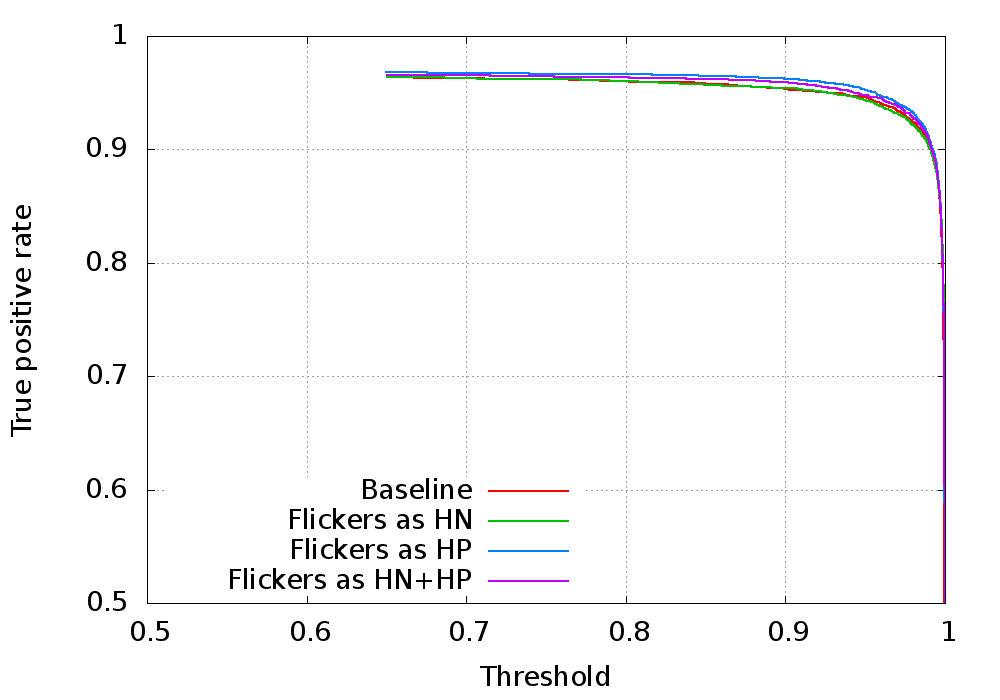}} \\
(a) & (b) & (c) \\
\end{tabular}
\caption{ Results on \textbf{FDDB}. (a) ROC curves comparing our hard example methods with the baseline Faster R-CNN detector; (b-c) separate plots showing False Positives and True Positive Rate with varying thresholds on detection score. }
\label{fig:fddb_roc}
\end{figure}

\noindent\textbf{Effect of Domain Shift on FDDB.}
The FDDB dataset~\cite{fddbTech} is comprised of 5,171 annotated faces in a set of 2,845 images taken from a subset of the Face in the Wild dataset. The images and the annotation style of FDDB have a significant \textit{domain shift} from WIDER Face, which are discussed in Jamal~et~al.~\cite{Jamal_2018_CVPR}. 
Fig.~\ref{fig:fddb_roc} compares our method with the Faster R-CNN baseline on FDDB, using the trained models from our experiments on WIDER Face (Sec.~\ref{sec:face_wider}). Although hard negatives reduce false positives (Fig.~\ref{fig:fddb_roc}(b)) and hard positives increase recall (Fig.~\ref{fig:fddb_roc}(c)), the performance does not consistently improve over the baseline on FDDB. We hypothesize that the large amounts of new training data result in shifting the original detector further away from the target FDDB domain, and this domain shift leads to a loss in performance. This may not have hurt our performance as much on WIDER Face because the domain shift between the relatively unconstrained WIDER images and our videos downloaded from YouTube was not severe enough to subsume the advantages from the hard examples. 

\noindent\textbf{Extension to Other Classes.}
The simplicity of our approach makes it easily extensible to other categories in a one-versus-rest setting. YouTube is a promising source of videos for various MS-COCO or PASCAL categories; mining hard negatives after that is fully automatic. To demonstrate this, we selected categories from MS-COCO and ran experiments to check if inclusion of hard negatives improves the baseline performance of a Faster R-CNN detector. We used the training method deployed by Sonntag~et~al.\cite{sonntag2017fine}, which allows for a convenient fine-tuning of the VGG16-based Faster R-CNN model on specific object classes of the MS-COCO dataset. The method was used to train a Faster R-CNN detector for a specific class vs background, starting from a multi-class VGG16 classifier pre-trained on Image-Net categories. This baseline detector was then used to mine hard negatives from downloaded YouTube videos of that category and then re-trained on the union of the new data and the original labeled training data. We show results for two categories: \textit{dogs} and \textit{trains}. A held out subset of the MS-COCO validation set was used for validating training hyper-parameters and the remainder of the validation data was used for evaluation.

For the \textit{dog} category, the labeled data was divided into train/val/test splits of 3041/177/1521 images. We manually selected and downloaded about 22 hours of dog videos from YouTube. 
We used the baseline dog detector to obtain detections on about 15 hours (1,296,000 frames at 24 fps) of dog videos. The hard negative mining algorithm was then run at a detector confidence threshold of 0.8. This yielded 2611 frames with at least one hard negative and one positive detection. The baseline model was then fine-tuned for 30k iterations on the union of the labeled MS-COCO data and the hard negatives. The hyper-parameters and best model were selected using a validation set. Similar experiments with \textit{trains} were performed, with train/val/test splits of 2464/157/1281 images. The results are summarized in the Table~\ref{tab:hn-dogs-trains}, where inclusion of hard negatives is observed to improve the baseline detector in both cases.
\begin{table}[ht]
\centering
\caption{Results on augmenting Faster R-CNN detectors with hard negatives for `\textit{dog}' and `\textit{train}' categories on MS-COCO. }
\label{tab:hn-dogs-trains}
\begin{tabular}{|l|l|l|l|l|l|}
\hline
\textbf{\begin{tabular}[c]{@{}l@{}}Category\end{tabular}} & \textbf{\begin{tabular}[c]{@{}l@{}}Model\end{tabular}} & \textbf{\begin{tabular}[c]{@{}l@{}}Training\\ iterations\end{tabular}} & \textbf{\begin{tabular}[c]{@{}l@{}}Training \\ hyperparams\end{tabular}}                         & \textbf{\begin{tabular}[c]{@{}l@{}}Validation\\ set AP\end{tabular}} & \textbf{\begin{tabular}[c]{@{}l@{}}Test\\ set AP\end{tabular}} \\ \hline
\multirow{2}{*}{\textbf{Dog}}                               & Baseline                                                         & 29000                                                                  & \begin{tabular}[c]{@{}l@{}}LR : 1e-3 for 10k,\\ 1e-4 for 10k-20k,\\ 1e-5 for 20k-29k\end{tabular} & 26.9                                                                 & 25.3                                                           \\ \cline{2-6} 
                                                                      & Flickers as HN                                                                   & 22000                                                                  & \begin{tabular}[c]{@{}l@{}}LR : 1e-4 for 15k,\\ 1e-5 for 15k-22k\end{tabular}                         & 28.1                                                                 & 26.4                                                  \\ \hline
\multirow{2}{*}{\textbf{Train}}                                & Baseline                                                      & 26000                                                                  & \begin{tabular}[c]{@{}l@{}}LR : 1e-3,\\ stepsize: 10k, \\ lr-decay: 0.1\end{tabular}                                                                                & 33.9                                                                 &  33.2                                                          \\ \cline{2-6} 
                                                                      & Flickers as HN                                                                   & 24000                                                                  & \begin{tabular}[c]{@{}l@{}}LR : 1e-3,\\ stepsize: 10k, \\ lr-decay: 0.1\end{tabular}                    & 35.4                                                                 & 33.7                                                  \\  \hline
\end{tabular}
\end{table}

\section{Conclusion}
\label{sec:conclusion}

This work leverages an existing phenomenon -- detector flicker in videos -- to mine hard negatives and hard positives at scale in an unsupervised manner. The usefulness of this method for improving an object detector is demonstrated on standard benchmarks for two well-known tasks -- face and pedestrian detection, using various detector architectures and supported by several ablation studies. The simplicity of our hard example mining approach makes it widely applicable to a variety of practical scenarios -- YouTube is a promising source of videos for almost any category and mining hard examples is a fully automatic procedure.  

\section*{Acknowledgment}
This research is based in part upon work supported by the Office of the Director of National Intelligence (ODNI), Intelligence Advanced Research Projects Activity (IARPA) under contract number 2014-14071600010 and in part on research sponsored by the Air Force Research Laboratory and DARPA under agreement number FA8750-18-2-0126. The views and conclusions contained herein are those of the authors and should not be interpreted as necessarily representing the official policies or endorsements, either expressed or implied, of ODNI, IARPA, the Air Force Research Laboratory and DARPA or the U.S. Government. The U.S. Government is authorized to reproduce and distribute reprints for Governmental purpose notwithstanding any copyright annotation thereon. 


\bibliographystyle{splncs04}
\bibliography{references}

\begin{thebibliography}{10}
\providecommand{\url}[1]{\texttt{#1}}
\providecommand{\urlprefix}{URL }
\providecommand{\doi}[1]{https://doi.org/#1}

\bibitem{Jamal_2018_CVPR}
Abdullah~Jamal, M., Li, H., Gong, B.: Deep face detector adaptation without
  negative transfer or catastrophic forgetting. In: The IEEE Conference on
  Computer Vision and Pattern Recognition (CVPR) (June 2018)

\bibitem{appel2013quickly}
Appel, R., Fuchs, T., Doll{\'a}r, P., Perona, P.: Quickly boosting decision
  trees--pruning underachieving features early. In: International Conference on
  Machine Learning. pp. 594--602 (2013)

\bibitem{athalye2017synthesizing}
Athalye, A., Sutskever, I.: Synthesizing robust adversarial examples. arXiv
  preprint arXiv:1707.07397  (2017)

\bibitem{blum1998combining}
Blum, A., Mitchell, T.: Combining labeled and unlabeled data with co-training.
  In: Proceedings of the eleventh annual conference on Computational learning
  theory. pp. 92--100. ACM (1998)

\bibitem{brazil2017illuminating}
Brazil, G., Yin, X., Liu, X.: Illuminating pedestrians via simultaneous
  detection \& segmentation. arXiv preprint arXiv:1706.08564  (2017)

\bibitem{cai2016unified}
Cai, Z., Fan, Q., Feris, R.S., Vasconcelos, N.: A unified multi-scale deep
  convolutional neural network for fast object detection. In: European
  Conference on Computer Vision. pp. 354--370. Springer (2016)

\bibitem{cai2015learning}
Cai, Z., Saberian, M., Vasconcelos, N.: Learning complexity-aware cascades for
  deep pedestrian detection. In: Proceedings of the IEEE International
  Conference on Computer Vision. pp. 3361--3369 (2015)

\bibitem{chang2017active}
Chang, H.S., Learned-Miller, E., McCallum, A.: Active bias: Training more
  accurate neural networks by emphasizing high variance samples. In: Advances
  in Neural Information Processing Systems. pp. 1003--1013 (2017)

\bibitem{chapelle2009semi}
Chapelle, O., Scholkopf, B., Zien, A.: Semi-supervised learning (chapelle, o.
  et al., eds.; 2006)[book reviews]. IEEE Transactions on Neural Networks
  \textbf{20}(3),  542--542 (2009)

\bibitem{dalal05histograms}
Dalal, N., Triggs, B.: Histograms of oriented gradients for human detection.
  In: CVPR. pp. 886--893 (2005). \doi{10.1109/CVPR.2005.177},
  \url{http://dx.doi.org/10.1109/CVPR.2005.177}

\bibitem{dollar2009integral}
Doll{\'a}r, P., Tu, Z., Perona, P., Belongie, S.: Integral channel features
  (2009)

\bibitem{dollar2009pedestrian}
Doll{\'a}r, P., Wojek, C., Schiele, B., Perona, P.: Pedestrian detection: A
  benchmark. In: Computer Vision and Pattern Recognition, 2009. CVPR 2009. IEEE
  Conference on. pp. 304--311. IEEE (2009)

\bibitem{dollar15fast}
Doll{\'{a}}r, P., Zitnick, C.L.: Fast edge detection using structured forests.
  {IEEE} Trans. Pattern Anal. Mach. Intell.  \textbf{37}(8),  1558--1570
  (2015). \doi{10.1109/TPAMI.2014.2377715},
  \url{http://dx.doi.org/10.1109/TPAMI.2014.2377715}

\bibitem{du2017fused}
Du, X., El-Khamy, M., Lee, J., Davis, L.: Fused dnn: A deep neural network
  fusion approach to fast and robust pedestrian detection. In: Applications of
  Computer Vision (WACV), 2017 IEEE Winter Conference on. pp. 953--961. IEEE
  (2017)

\bibitem{farfade15multi}
Farfade, S.S., Saberian, M.J., Li, L.: Multi-view face detection using deep
  convolutional neural networks. In: ICMR. pp. 643--650 (2015).
  \doi{10.1145/2671188.2749408},
  \url{http://doi.acm.org/10.1145/2671188.2749408}

\bibitem{felzenszwalb2010object}
Felzenszwalb, P.F., Girshick, R.B., McAllester, D., Ramanan, D.: Object
  detection with discriminatively trained part-based models. IEEE transactions
  on pattern analysis and machine intelligence  \textbf{32}(9),  1627--1645
  (2010)

\bibitem{friedman2000additive}
Friedman, J., Hastie, T., Tibshirani, R., et~al.: Additive logistic regression:
  a statistical view of boosting (with discussion and a rejoinder by the
  authors). The annals of statistics  \textbf{28}(2),  337--407 (2000)

\bibitem{geman1986markov}
Geman, S., Graffigne, C.: Markov random field image models and their
  applications to computer vision. In: Proceedings of the international
  congress of mathematicians. vol.~1, p.~2 (1986)

\bibitem{girshick15fast}
Girshick, R.B.: Fast {R-CNN}. In: ICCV. pp. 1440--1448 (2015).
  \doi{10.1109/ICCV.2015.169}, \url{http://dx.doi.org/10.1109/ICCV.2015.169}

\bibitem{girshick14rich}
Girshick, R.B., Donahue, J., Darrell, T., Malik, J.: Rich feature hierarchies
  for accurate object detection and semantic segmentation. In: CVPR. pp.
  580--587 (2014). \doi{10.1109/CVPR.2014.81},
  \url{http://dx.doi.org/10.1109/CVPR.2014.81}

\bibitem{he14spatial}
He, K., Zhang, X., Ren, S., Sun, J.: Spatial pyramid pooling in deep
  convolutional networks for visual recognition. In: ECCV. pp. 346--361 (2014)

\bibitem{hosang2015taking}
Hosang, J., Omran, M., Benenson, R., Schiele, B.: Taking a deeper look at
  pedestrians. In: Proceedings of the IEEE Conference on Computer Vision and
  Pattern Recognition. pp. 4073--4082 (2015)

\bibitem{hu2017finding}
Hu, P., Ramanan, D.: Finding tiny faces. In: 2017 IEEE Conference on Computer
  Vision and Pattern Recognition (CVPR). pp. 1522--1530. IEEE (2017)

\bibitem{fddbTech}
Jain, V., Learned-Miller, E.: {FDDB}: A benchmark for face detection in
  unconstrained settings. Tech. Rep. UM-CS-2010-009, University of
  Massachusetts, Amherst (2010)

\bibitem{jiang2017face}
Jiang, H., Learned-Miller, E.: Face detection with the faster r-cnn. In:
  Automatic Face \& Gesture Recognition (FG 2017), 2017 12th IEEE International
  Conference on. pp. 650--657. IEEE (2017)

\bibitem{erdosrenyi}
Jin, S., Su, H., Stauffer, C., Learned-Miller, E.: End-to-end face detection
  and cast grouping in movies using erdos-renyi clustering. In: ICCV (2017)

\bibitem{kalal2010pn}
Kalal, Z., Matas, J., Mikolajczyk, K.: Pn learning: Bootstrapping binary
  classifiers by structural constraints. In: Computer Vision and Pattern
  Recognition (CVPR), 2010 IEEE Conference on. pp. 49--56. IEEE (2010)

\bibitem{klaser2010human}
Kl{\"a}ser, A., Marsza{\l}ek, M., Schmid, C., Zisserman, A.: Human focused
  action localization in video. In: European Conference on Computer Vision. pp.
  219--233. Springer (2010)

\bibitem{li15a}
Li, H., Lin, Z., Shen, X., Brandt, J., Hua, G.: A convolutional neural network
  cascade for face detection. In: CVPR. pp. 5325--5334 (2015).
  \doi{10.1109/CVPR.2015.7299170},
  \url{http://dx.doi.org/10.1109/CVPR.2015.7299170}

\bibitem{li2017scale}
Li, J., Liang, X., Shen, S., Xu, T., Feng, J., Yan, S.: Scale-aware fast r-cnn
  for pedestrian detection. IEEE Transactions on Multimedia  (2017)

\bibitem{li16face}
Li, Y., Sun, B., Wu, T., Wang, Y., Gao, W.: Face detection with end-to-end
  integration of a convnet and a 3d model. ECCV  \textbf{abs/1606.00850}
  (2016),
  \url{http://dblp.uni-trier.de/db/journals/corr/corr1606.html#LiSWW016}

\bibitem{lin2017feature}
Lin, T.Y., Doll{\'a}r, P., Girshick, R., He, K., Hariharan, B., Belongie, S.:
  Feature pyramid networks for object detection. In: CVPR. vol.~1, p.~4 (2017)

\bibitem{lin2017focal}
Lin, T.Y., Goyal, P., Girshick, R., He, K., Doll{\'a}r, P.: Focal loss for
  dense object detection. arXiv preprint arXiv:1708.02002  (2017)

\bibitem{liu2016ssd}
Liu, W., Anguelov, D., Erhan, D., Szegedy, C., Reed, S., Fu, C.Y., Berg, A.C.:
  Ssd: Single shot multibox detector. In: European conference on computer
  vision. pp. 21--37. Springer (2016)

\bibitem{loshchilov2015online}
Loshchilov, I., Hutter, F.: Online batch selection for faster training of
  neural networks. arXiv preprint arXiv:1511.06343  (2015)

\bibitem{lu2017no}
Lu, J., Sibai, H., Fabry, E., Forsyth, D.: No need to worry about adversarial
  examples in object detection in autonomous vehicles. arXiv preprint
  arXiv:1707.03501  (2017)

\bibitem{luo2015foveation}
Luo, Y., Boix, X., Roig, G., Poggio, T., Zhao, Q.: Foveation-based mechanisms
  alleviate adversarial examples. arXiv preprint arXiv:1511.06292  (2015)

\bibitem{ozerov2013evaluating}
Ozerov, A., Vigouroux, J.R., Chevallier, L., P{\'e}rez, P.: On evaluating face
  tracks in movies. In: Image Processing (ICIP), 2013 20th IEEE International
  Conference on. pp. 3003--3007. IEEE (2013)

\bibitem{ranjan15a}
Ranjan, R., Patel, V.M., Chellappa, R.: A deep pyramid deformable part model
  for face detection. In: BTAS. pp.~1--8. IEEE (2015),
  \url{http://dblp.uni-trier.de/db/conf/btas/btas2015.html#RanjanPC15}

\bibitem{redmon2016you}
Redmon, J., Divvala, S., Girshick, R., Farhadi, A.: You only look once:
  Unified, real-time object detection. In: Proceedings of the IEEE conference
  on computer vision and pattern recognition. pp. 779--788 (2016)

\bibitem{ren16faster}
Ren, S., He, K., Girshick, R., Sun, J.: Faster {R-CNN}: Towards real-time
  object detection with region proposal networks. {IEEE} Trans. Pattern Anal.
  Mach. Intell.  (2016)

\bibitem{ren15faster}
Ren, S., He, K., Girshick, R.B., Sun, J.: Faster {R-CNN:} towards real-time
  object detection with region proposal networks. In: NIPS. pp. 91--99 (2015),
  \url{http://papers.nips.cc/paper/5638-faster-r-cnn-towards-real-time-object-detection-with-region-proposal-networks}

\bibitem{rosenberg2005semi}
Rosenberg, C., Hebert, M., Schneiderman, H.: Semi-supervised self-training of
  object detection models  (2005)

\bibitem{rowley1998neural}
Rowley, H.A., Baluja, S., Kanade, T.: Neural network-based face detection. IEEE
  Transactions on pattern analysis and machine intelligence  \textbf{20}(1),
  23--38 (1998)

\bibitem{scholkopf2002learning}
Sch{\"o}lkopf, B., Smola, A.J.: Learning with kernels: support vector machines,
  regularization, optimization, and beyond. MIT press (2002)

\bibitem{shrivastava2016training}
Shrivastava, A., Gupta, A., Girshick, R.: Training region-based object
  detectors with online hard example mining. In: Proceedings of the IEEE
  Conference on Computer Vision and Pattern Recognition. pp. 761--769 (2016)

\bibitem{simo2014fracking}
Simo-Serra, E., Trulls, E., Ferraz, L., Kokkinos, I., Moreno-Noguer, F.:
  Fracking deep convolutional image descriptors. CoRR, abs/1412.6537
  \textbf{2} (2014)

\bibitem{singh2016track}
Singh, K.K., Xiao, F., Lee, Y.J.: Track and transfer: Watching videos to
  simulate strong human supervision for weakly-supervised object detection. In:
  CVPR. vol.~1, p.~2 (2016)

\bibitem{sonntag2017fine}
Sonntag, D., Barz, M., Zacharias, J., Stauden, S., Rahmani, V., F{\'o}thi,
  {\'A}., L{\H{o}}rincz, A.: Fine-tuning deep cnn models on specific ms coco
  categories. arXiv preprint arXiv:1709.01476  (2017)

\bibitem{stalder2010cascaded}
Stalder, S., Grabner, H., Van~Gool, L.: Cascaded confidence filtering for
  improved tracking-by-detection. In: European Conference on Computer Vision.
  pp. 369--382. Springer (2010)

\bibitem{sun2017face}
Sun, X., Wu, P., Hoi, S.C.: Face detection using deep learning: An improved
  faster rcnn approach. arXiv preprint arXiv:1701.08289  (2017)

\bibitem{sung1994learning}
Sung, K.K., Poggio, T.: Learning and example selection for object and pattern
  detection  (1994)

\bibitem{sutton2006introduction}
Sutton, C., McCallum, A.: An introduction to conditional random fields for
  relational learning, vol.~2. Introduction to statistical relational learning.
  MIT Press (2006)

\bibitem{tang2012shifting}
Tang, K., Ramanathan, V., Fei-Fei, L., Koller, D.: Shifting weights: Adapting
  object detectors from image to video. In: Advances in Neural Information
  Processing Systems. pp. 638--646 (2012)

\bibitem{wan2016bootstrapping}
Wan, S., Chen, Z., Zhang, T., Zhang, B., Wong, K.k.: Bootstrapping face
  detection with hard negative examples. arXiv preprint arXiv:1608.02236
  (2016)

\bibitem{wang2017fast}
Wang, X., Shrivastava, A., Gupta, A.: A-fast-rcnn: Hard positive generation via
  adversary for object detection  (2017)

\bibitem{wang2017detecting}
Wang, Y., Ji, X., Zhou, Z., Wang, H., Li, Z.: Detecting faces using
  region-based fully convolutional networks. arXiv preprint arXiv:1709.05256
  (2017)

\bibitem{westonlarge}
WESTON, J.: Large-scale semi-supervised learning

\bibitem{yang2012online}
Yang, B., Nevatia, R.: An online learned crf model for multi-target tracking.
  In: Computer Vision and Pattern Recognition (CVPR), 2012 IEEE Conference on.
  pp. 2034--2041. IEEE (2012)

\bibitem{yang15from}
Yang, S., Luo, P., Loy, C.C., Tang, X.: From facial parts responses to face
  detection: {A} deep learning approach. In: ICCV. pp. 3676--3684 (2015).
  \doi{10.1109/ICCV.2015.419}, \url{http://dx.doi.org/10.1109/ICCV.2015.419}

\bibitem{yang16wider}
Yang, S., Luo, P., Loy, C.C., Tang, X.: {WIDER FACE}: A face detection
  benchmark. In: CVPR (2016)

\bibitem{yu2016unitbox}
Yu, J., Jiang, Y., Wang, Z., Cao, Z., Huang, T.: Unitbox: An advanced object
  detection network. In: Proceedings of the 2016 ACM on Multimedia Conference.
  pp. 516--520. ACM (2016)

\bibitem{zhang2016joint}
Zhang, K., Zhang, Z., Li, Z., Qiao, Y.: Joint face detection and alignment
  using multitask cascaded convolutional networks. IEEE Signal Processing
  Letters  \textbf{23}(10),  1499--1503 (2016)

\bibitem{zhang2016faster}
Zhang, L., Lin, L., Liang, X., He, K.: Is faster r-cnn doing well for
  pedestrian detection? In: European Conference on Computer Vision. pp.
  443--457. Springer (2016)

\bibitem{zhang2016far}
Zhang, S., Benenson, R., Omran, M., Hosang, J., Schiele, B.: How far are we
  from solving pedestrian detection? In: Proceedings of the IEEE Conference on
  Computer Vision and Pattern Recognition. pp. 1259--1267 (2016)

\bibitem{zhang2017s}
Zhang, S., Zhu, X., Lei, Z., Shi, H., Wang, X., Li, S.Z.: S3fd: Single shot
  scale-invariant face detector. arXiv preprint arXiv:1708.05237  (2017)

\bibitem{zitnick2014edge}
Zitnick, C.L., Doll{\'a}r, P.: Edge boxes: Locating object proposals from
  edges. In: European Conference on Computer Vision. pp. 391--405. Springer
  (2014)

\end{thebibliography}
\end{document}